\documentclass[10.5pt]{article}
\usepackage[utf8]{inputenc}
\usepackage{amsmath}

\usepackage{geometry}
\usepackage[hidelinks]{hyperref}
\hypersetup{bookmarksnumbered}

\usepackage[nodisplayskipstretch]{setspace}

\usepackage[hang, flushmargin, bottom]{footmisc}
\usepackage{xcolor}

\usepackage{natbib}
\usepackage{enumitem}
\setlist{nolistsep}

\usepackage{amsmath, amssymb, bm, mathtools, nccmath, mathtools, esint, fmtcount}

\usepackage{calc}
\usepackage{float}

\usepackage{array, multirow, multicol, tabularx}
\newcolumntype{Y}{>{\centering\arraybackslash}X}
\newcolumntype{L}[1]{>{\raggedright\let\newline\\\arraybackslash\hspace{0pt}}m{#1}}
\newcolumntype{C}[1]{>{\centering\let\newline\\\arraybackslash\hspace{0pt}}m{#1}}
\newcolumntype{R}[1]{>{\raggedleft\let\newline\\\arraybackslash\hspace{0pt}}m{#1}}

\usepackage[font=footnotesize, labelfont=bf]{caption}
\usepackage{subcaption, graphicx, textcomp, xurl}           
\usepackage{titlesec}
\titleformat{\section}
{\normalfont\fontsize{14}{16.8}\bfseries}{\thesection}{1em}{}
\titlespacing\section{0pt}{12pt plus 4pt minus 2pt}{6pt}

\titleformat{\subsection}
{\normalfont\fontsize{11}{13.2}\bfseries}{\thesubsection}{1em}{}
\titlespacing\subsection{0pt}{11pt plus 2pt minus 2pt}{4pt plus 2pt minus 2pt}

\titleformat{\subsubsection}
{\normalfont\fontsize{11}{13.2}\itshape}{\thesubsubsection}{1em}{}
\titlespacing\subsubsection{0pt}{11pt plus 2pt minus 2pt}{4pt plus 2pt minus 2pt}           
\begin{document}

\title{Simulation-Driven Reinforcement Learning in Queuing Network Routing Optimization}

\author{
  Fatima Al-Ani\thanks{These authors contributed equally.},
  Molly Wang\footnotemark[1],
  Jevon Charles\footnotemark[1],\\
  Aaron Ong,
  Joshua Forday,
  Vinayak Modi\thanks{All authors were Master’s students at the Department of Computing, Imperial College London when this paper was written. They graduated in August 2024.}\\
  {\small Department of Computing, Imperial College London, UK}
}

\maketitle

\begin{abstract}
This report details the development of a simulation-driven reinforcement learning (RL) framework for optimizing routing decisions in complex queueing network systems, with a particular emphasis on manufacturing and communication applications. Recognizing the limitations of traditional queueing methods—which often struggle with dynamic, uncertain environments—we propose a robust RL approach leveraging Deep Deterministic Policy Gradient (DDPG) combined with Dyna-style planning (Dyna-DDPG). The framework includes a flexible and configurable simulation environment capable of modeling diverse queueing scenarios, disruptions, and unpredictable conditions. Our enhanced Dyna-DDPG implementation incorporates separate predictive models for next-state transitions and rewards, significantly improving stability and sample efficiency. Comprehensive experiments and rigorous evaluations demonstrate the framework's capability to rapidly learn effective routing policies that maintain robust performance under disruptions and scale effectively to larger network sizes. Additionally, we highlight strong software engineering practices employed to ensure reproducibility and maintainability of the framework, enabling practical deployment in real-world scenarios.
\end{abstract}


\section*{Introduction}
The aim of this project was to develop a simulation-based RL framework to address optimisation problems in practical network-based systems. The package was designed in collaboration with Datasparq, a London-based data-driven technology firm specialising in artificial intelligence solutions for real-world problems. The particular use setting focused on industrial manufacturing systems; however, the package can be extended to support other systems that can be modelled as queuing networks. Below, we explain the motivations for choosing these particular project settings.

\bigskip
\textbf{Queue Network-based Modelling}. Queues are a fundamental aspect of many physical systems, including telecommunications networks, computer systems, and manufacturing processes. The understanding and optimisation of queuing networks enhance resource allocation, enable system performance analysis, and reduce operational costs for organisations. However, queuing networks often involve complex interactions, such as multiple interconnected queues and stochastic processes in network traffic. Traditional optimisation methods, such as Matrix-analytic methods and Birth-Death Processes, tend to offer limited state representations and primarily focus on single-queue scenarios. Additionally, they lack the flexibility required for real-time adaptation, a crucial need in queuing networks that must dynamically adjust to fluctuations in demand.

\bigskip
\textbf{Reinforcement Learning-Powered Approach}. Reinforcement learning (RL) offers a powerful solution to the aforementioned limitations. RL algorithms naturally incorporate stochasticity during their training processes; for instance, the Temporal-Difference learning method continually updates its value estimations based on new experiences, making it ideally suited for stochastic settings. Moreover, RL employs the well-known exploration-and-exploitation paradigm, allowing the agent to explore the environment by trying different actions and observing outcomes, thereby inherently managing variability.

\bigskip
\textbf{Motivation for Continuous RL}. Although discrete reinforcement learning allows for a greater explanation of the agent's behaviours, the curse of dimensionality poses a significant challenge. As the state space expands, the application and scalability of the discrete models become severely limited. Therefore, in this project, we adopted a continuous RL approach to manage high-dimensional environments, which are representative of real-world systems.

\bigskip
\textbf{Simulation-Driven Approach}. From a theoretical standpoint, RL algorithms should ideally be trained in live systems, as obtaining real experience would enable ground-truth learning. However, mistakes made during live training can be costly and may lead to dangerous situations—particularly within the given context of manufacturing systems. As such, we use simulation-driven RL to enable safe testing and thorough evaluation of the learned policy without the risk of human harm and equipment or environmental damage.

\bigskip
The existing landscape of deep RL libraries is not equipped with the functionality required for this setting. The current offerings lack support for highly customisable queuing networks, do not provide a solid framework for evaluating policy or agent decisions, and require a substantial understanding of deep learning packages for agent configuration. These challenges complicate the implementation process and detract from the time that could be better spent on analysing the agent’s performance and refining its capabilities. Taking these requirements into account, our library was designed to provide the following functionalities:

\bigskip
\begin{itemize}

    \item Configurable queuing environments for the modelling of manufacturing systems. These are extensible to other systems which can be modelled effectively as queuing networks.

    \bigskip
    \item A specialised RL agent that is capable of managing continuous action spaces, particularly suited for graph-style topologies.

    \bigskip
    \item An end-to-end training, tuning, and evaluation suite that simplifies the intricacies of deep RL training, allowing stakeholders to focus on higher-level strategy and analysis.

    \bigskip
    \item An extensive evaluation framework dedicated to the thorough investigation of agent training, learning behaviours, and policy assessment, ensuring that each aspect of the RL process is analysable.
\end{itemize}

\section*{Background}

\subsection{Queueing Networks}
\subsubsection{Overview of queueing theory}

Queueing theory \citep{sundarapandian2009queueing} is a field of operational research involving the mathematical study of waiting lines and queues, particularly focusing on the formation, function, and congestion of these lines. Central to queueing theory is the modelling of how items, referred to as jobs, move through a network of limited resources that are shared among competing demanding entities. Queueing networks are defined by the following components:
\begin{itemize}
    \item The arrival process of the job, which describes how jobs enter a queue. The arrival process is often modelled by statistical distributions such as the Poisson distribution.
    \item The service times, which are the duration of time required to serve a job. In practice, this is often modelled by exponential distributions which indicate variable service durations.
    \item The number of servers or nodes in the network, which refers to the individual points within the queue system where services are provided. Each server handles incoming tasks or requests from the jobs based on the service process.
    \item The buffer size at each queue, which represents the maximum number of jobs allowed in the queue.
    \item The queue discipline, which is the order in which jobs are served. Ubiquitous disciplines include FIFO (First In First Out) and LIFO (Last In First Out).
\end{itemize}

\bigskip
The foundational model in queueing theory is the M/M/1 model \citep{sundarapandian2009queueing}, which is characterised by Markovian arrivals, Markovian service processes, and a single server. Analysis of queueing network components provides critical performance metrics such as the average waiting time, system utilisation, and the throughput rate—the rate at which jobs complete processes and exit the network. Thus, queueing theory is employed in many fields to optimise resource allocation and enhance service efficiency.

\subsubsection{Queueing networks used as manufacturing systems modellers}
In manufacturing systems, elements such as products, materials, and tasks flow through the system with a degree of uncertainty, causing delays and underutilisation of resources \citep{Saini2024Queueing}. Queueing models, specifically the classical Jackson network, \citep{Buzacott1986QueueingNM} have been extensively used in manufacturing optimisation to address these challenges by providing a structured way to analyse and optimise the flow of elements and resource allocation. The Jackson network can be described as an open queueing model which accepts external arrivals. It consists of a defined number of servers, where each server is a simple M/M/1 queue. This network is defined as an open model, due to the fact that jobs arrive from outside the network at arrival times following a Poisson distribution with respect to a defined arrival rate. The processing time at each server—the service time—is exponential with respect to a defined service rate. Jobs routing from one server to another follow a Markovian chain governed by the assigned routing probabilities, also called the transition probabilities. These assumptions establish the Jackson model as a continuous-time Markov chain, owing to the memoryless property of the exponential distribution, which reinforces the Markov property: the future state of the system depends only on the current state, not on the path by which it arrived there.

\bigskip
\citeauthor{Chen2018} utilised queueing networks in manufacturing optimisation, where a modular Automated Guided Vehicle-based manufacturing system was conceptualised as an open queueing network. In this model, the flow of units was simulated through various workstations. Each assembly point was modelled as a node in the network, and the units’ flow paths were defined as the queues. The study demonstrated that the queueing model simulation was effective in identifying bottlenecks, determining scheduling strategies that minimise time and enhance flow, and recommending operational strategies that achieve the highest throughput rates.

\subsection{Reinforcement Learning}
\subsubsection{Simulation-Driven RL}
Reinforcement Learning is a method that uses a trial-and-error approach to identify the appropriate action to take in a given state within an environment. Training an RL agent directly in real-world systems can often be inefficient, impractical, and potentially unsafe due to the instability of policies during the learning phase. To mitigate these risks, simulated replicas of physical environments are used as safe and cost-effective alternatives for evaluating agent performance prior to deployment. These simulations not only allow for the testing of policies under controlled conditions but also enable the creation of diverse scenarios that might be difficult to replicate in real life. This helps ensure that the learned policies can generalise effectively across various potential situations. In the context of route optimisation, for example, agents learn to make optimal decisions even when faced with dysfunctional nodes or servers and complex rerouting loops. Additionally, the effectiveness of these policies can be assessed through live simulations, allowing for a comprehensive evaluation of their quality before actual implementation.

\subsubsection{Motivation for Continuous RL}
The successful adoption of RL-based systems in real-world applications relies heavily on the explainability of the agent’s policies. In industrial settings, discrete RL agents are often preferred for their tabular value functions, which facilitate clear explanations and evaluations of decisions. However, route optimisation poses specific challenges related to the discretisation of the action space. Excessively fine discretisation leads to an unmanageable state space, while overly coarse discretisation results in a controller that is not practical. To overcome these challenges, we adopted the DDPG (Deep Deterministic Policy Gradient) agent, which operates within a continuous action space. This configuration allows for more precise control over routing proportions and enables the agent to not only select the most reasonable actions on average but also to explore actions at the extremes of this distribution. Such flexibility significantly enhances the agent’s ability to adapt to and effectively manage the complexities of route optimisation, ensuring a comprehensive and robust approach.

\subsubsection{Dyna-DDPG}
\label{subsection:dyna-ddpg}
Deep Deterministic Policy Gradient (DDPG) introduced by \citeauthor{lillicrap2015continuous} is a model-free approach to reinforcement learning problems with a continuous action space, such as the queueing system problem under consideration. The agent directly learns the optimal policy $\pi(s)$ by computing the gradient of the expected return with respect to the policy parameters $\theta$. Similar to the well-known Deep Q-Network (DQN) algorithm \citep{Mnih2015HumanlevelCT}, DDPG also uses experience replay to break correlations between consecutive samples, making them more independent and identically distributed. The DDPG algorithm has two main disadvantages: (i) it requires a large number of samples to learn efficiently, leading to long training times, and (ii) it can suffer from insufficient exploration of peripheral states in high-dimensional systems.

\bigskip
Dyna-DDPG \citep{zhang2020deep} adds elements of Dyna \citep{dyna1991}, a model-based algorithm, to DDPG. Dyna incorporates planning into reinforcement learning by keeping an internal model of the environment within the agent. Thus, the agent can internally predict the next state and reward given a state-action pair as input. This internal model is gradually improved during training as the agent collects experiences within the real environment. \textit{Planning} involves using the agent's internal model to simulate or `hallucinate' additional experiences, which are used to update the network's parameters. The planning step increases both sample efficiency and the number of states explored, which, in theory, addresses both shortcomings of DDPG simultaneously. However, planning requires additional compute for each experience sampled from the replay buffer, and this is significant for high-dimensional systems. Yet, we demonstrate in \textit{section} \ref{section:analyis-and-evaluation} that the algorithm's training time scales well with larger queueing networks.

\section*{Software Design and Development}

The task of developing an RL agent to optimise a queueing network was strategically divided into two main phases. In the first phase, decisions were made concerning the core RL problem specification. These included defining the state and action spaces as well as designing the reward function. The second phase focused on the implementation of the RL architecture, involving the design of the RL agent itself and its simulation environment. After the major architectural decisions were made and implemented, the required components of the training, tuning and evaluation pipeline were constructed.

\bigskip
In the following sections, we will elaborate on the design decisions made during the development process. Additionally, we will provide an overview of the evaluation functions used to assess various dimensions of the agent’s performance, with a detailed analysis available in section~\ref{section:analyis-and-evaluation}. Finally, we will discuss the operational rules and software engineering processes implemented to deliver our final package. Instructions for installing and using the code are available in Appendix B (page \pageref{appendix:installation and usage}) and on GitHub's README.

\subsection{Framing the Reinforcement Learning Problem}
\subsubsection{State Representation}
After multiple attempts to represent the RL agent's state space, we concluded by defining the average end-to-end delay for each node as the state. This ensured the agent perceived all the relevant and representative information about the queueing environment required for informed decision making. It is computed as the average time difference between the job's arrival and departure from each node for all jobs travelling through that node. In our repository, the function \texttt{get\_state()} of the \texttt{rl\_env} module is used to retrieve the current state of the environment, which is represented as an array where each element corresponds to the average end-to-end delay for each queue in the network. To mathematically represent the state, we:

\begin{enumerate}
    \item Retrieve the queue data for each edge \( i \) as \( Q_i \). If the queue data for a packet indicates it has not exited yet (end time \( = 0 \)), update this to the current time.
    \item Calculate the end-to-end delay \( D_{i,j} \) for each packet \( j \) in \( Q_i \) as:
    \[
    D_{i,j} = \text{exit time} - \text{arrival time}
    \]
    \item Compute the total end-to-end delay for edge \( i \) as:
    \[
    \text{Total EtE Delay}_i = \sum_{j} D_{i,j}
    \]
    \item If there are jobs that have passed through the queue (i.e., \( Q_i \) is not empty), calculate the average end-to-end delay for edge \( i \) by dividing the total delay by the number of jobs:
    \[
    \overline{D}_i = \frac{\text{Total EtE Delay}_i}{\text{number of jobs in } Q_i}
    \]
    \item If no jobs have passed through the queue, set the average end-to-end delay for that edge to zero.
\end{enumerate}

\bigskip
Each \( \overline{D}_i \) is then included in the state array returned by the function. The state array provides a snapshot of the network's performance in terms of delays across all active edges with a \texttt{LossQueue}.

\bigskip
An example of a five-node queueing network's state is shown in the table below:

\begin{table}[h]
\centering
\begin{tabular}{|l|c|c|c|c|c|}
\hline
\textbf{Node ID} & Node 1 & Node 2 & Node 3 & Node 4 & Node 5 \\ \hline
\textbf{EoE Delay} & 1.46 & 51.01 & 1.01 & 67.12 & 3.72 \\ \hline
\end{tabular}
\caption{An example of average end-to-end delay for each node}
\label{tab:eoe_delay}
\end{table}

As we can see, it identifies which nodes are experiencing longer delays and which are performing normally. The longer the end-to-end delays for specific nodes, as indicated in the above example for nodes 2 and 4, the more urgently this network requires actions from the agent to redistribute workloads around these nodes.

\subsubsection{Action Representation and Its Mapping to Network Routing}
As mentioned previously, the original action space took a discrete tabular approach which provided an advantage in simplicity. However, this approach resulted in a drawback that only discrete actions could be taken by the agent. However, the required granularity of routing proportions could be vastly different from one queueing system to another. Moreover, the output of the agent was inherently deterministic which limits the exploration space. Therefore, the most optimal solution may not be found. Furthermore, due to the inherent nature of the queueing network, one action will have a compounding effect on the consequent queues and thus a continuous action space allows for simultaneous adjustments to multiple queues to observe the action. \citep{shen2020deep}. Hence, to allow our agent to effectively impact the network, we employed a vector of values as our action ranging from 0 to 1 to represent the probability for each queue in the network. This was achieved using \texttt{select\_action()} function under \texttt{ddpg\_agent.py}. This function selects an action based on the current state of the queueing environment using a neural network, \texttt{Actor}. The process is mathematically represented as follows:

\bigskip
\begin{enumerate}
    \item The state of the environment, denoted as \( s_t \), is passed to the function. The neural network, referred to as the \texttt{actor}, takes \( s_t \) as input.

    \bigskip
    \item Convert the state \( s_t \) into a suitable format (e.g., a floating point tensor) for neural network processing:
    \[
    s_t^{\text{float}} = \text{float}(s_t)
    \]
    \item The neural network processes the input state and outputs the action \( a_t \) as:
    \[
    a_t = \text{actor}(s_t^{\text{float}})
    \]
    \item The output \( a_t \) is then detached from the current computation graph to prevent further gradient calculations:
    \[
    a_t = a_t.\text{detach}(\;)
    \]
    \item The action \( a_t \) is thus selected and ready for application in the environment.
\end{enumerate}

\bigskip
This method demonstrates the decision-making capability of the neural network where its architecture and weights determine how the state is transformed into an action. The neural network serves as a function approximator, mapping states to actions in a manner optimised through training.

\bigskip
The next step is to convert these weights to routing probabilities at different nodes during decision-making. This linking is accomplished via a masking layer, which will be detailed shortly. An example of the action vector is shown in the table below:

\begin{table}[!h]
\centering
\begin{tabular}{|l|c|c|c|c|c|}
\hline
\textbf{Queue ID} & Queue 1 & Queue 2 & Queue 3 & Queue 4 & Queue 5 \\ \hline
\textbf{Probability} & 0.71 & 0.27 & 0.30 & 0.59 & 0.83 \\ \hline
\end{tabular}
\caption{Probabilities for each queue in the action vector}
\label{tab:action_vector}
\end{table}

\bigskip
As we can see, the higher the likelihood for a specific queue, the more favourable it is to route jobs to that queue. Also, queues with greater probability process jobs faster than others, therefore allocating jobs to this queue can help reduce overall network traffic.

\bigskip
However, the queueing network operates by routing jobs between nodes. Thus, the action vector must be transformed into routing probabilities to convert abstract actions (learned weights) into tangible decisions (task routing via the network).

\bigskip
This transformation is achieved through a masking layer which utilises the variable \texttt{edge\_list}, a dictionary that shows how nodes are connected to each other through different queues. For example, in the \texttt{edge\_list} shown below for our queue network,

\begin{figure}[h]
\begin{align*}
&\text{env.qn\_net.edge\_list} \\
&\text{\{ 0: \{1: 1\},   1: \{2: 2, 3: 3, 4: 4\},   2: \{5: 5\},   3: \{6: 6, 7: 7\},   4: \{8: 8\},} \\
&\text{5: \{9: 9\},   6: \{9: 10\},   7: \{9: 11\},   8: \{9: 12\},   9: \{10: 0\} \}} \\[-2em]
\end{align*}
\caption{Example edge list}
\label{fig:edge list}
\end{figure}

We can see that the queue type (or edge type) that is connecting node 0 and node 1 is Type 1. Additionally, the queue types that are connected to node 1 are Type 2, 3, and 4 which links to nodes 2, 3, 4 accordingly. Take node 1 for instance, to extract its routing probabilities from the action vector, we would extract the 2\textsuperscript{nd}, 3\textsuperscript{rd}, 4\textsuperscript{th} elements in the vector. This conversion is achieved in \texttt{get\_next\_state()} function under the RL environment file.

\bigskip
As a result, by extracting elements for each node in the action vector by their queue types, we can retrieve their routing probabilities. In this way, it is more interpretable and allows users to understand what the agent is doing and why.

\subsubsection{Reward Function}
To appropriately evaluate the agent's behaviours, we employ a reward function that considers both server delays and system throughput. The reward is calculated by taking the average end-to-end delays for all nodes and dividing by the throughput ratio. The throughput ratio is defined as the number of jobs that arrive at the network divided by the number of jobs that leave the network. Mathematically, let \( E \) be the set of indices for edges in the network where each edge corresponds to a \texttt{LossQueue}. For each edge \( i \in E \), define:
\begin{itemize}
    \item \( Q_i \) as the set of queue data for edge \( i \) after a specified number of entries.
    \item \( S_i \) as the set of indices in \( Q_i \) where the packet was serviced (not lost).
    \item \( T_i = |S_i| \) as the throughput for edge \( i \), i.e., the count of serviced jobs.
    \item \( D_{i,j} \) as the end-to-end delay for packet \( j \) in \( S_i \), calculated as \( \text{service time} - \text{arrival time} \).
\end{itemize}

\bigskip
The average delay for edge \( i \) is given by:
\[
\overline{D}_i = \frac{1}{T_i} \sum_{j \in S_i} D_{i,j}
\]
The overall average delay across all such edges is:
\[
\overline{D} = \frac{1}{|E|} \sum_{i in E} \overline{D}_i
\]
Define \( E' \) as the set of edges corresponding to \texttt{NullQueue}, and \( E'' \) as the set of edges with arrivals.
\begin{itemize}
    \item Let \( X_i \) be the number of exits (jobs exiting the queue) for each \( i in E' \).
    \item Let \( A_i \) be the number of arrivals for each \( i in E'' \).
\end{itemize}

\bigskip
The throughput ratio \( R \) is calculated as:
\[
R = \frac{\sum_{i in E'} X_i}{\sum_{i in E''} A_i}
\]

The reward \( \mathcal{R} \) is given by:
\[
\mathcal{R} = -\frac{\overline{D}}{R}
\]
This formula implies that the reward is negatively influenced by the average delay and inversely related to the throughput ratio. Lower delays and higher throughput ratios are preferable, leading to a higher reward.

\bigskip
In real-world circumstances, network delays, particularly those involving telecommunications or data routing, can have a direct impact on user experiences and system efficiency. Thus, by inversely linking the reward to average delays, the function motivates the agent to seek out strategies that reduce delays. Throughput, on the other hand, refers to the rate at which jobs are performed. By encouraging higher throughput, the agent can significantly boost the flow of processed jobs.

\subsection{Reinforcement Learning Training Architecture}
\subsubsection{Agent}
This module encompasses the required components to train RL agent using the Deep Deterministic Policy Gradient (DDPG) algorithm. The associated components can be broken down as follows:

\begin{itemize}
    \item \textbf{\texttt{model.py:}} Contains the neural network architectures for the four unique neural networks that are part of the RL agent: the actor network, critic network, reward predictor network, and next state predictor network.
    
    A decision was made to split the the original Dyna-DDPG algorithm's, single neural network which predicts both next state and reward given a state-action pair into two smaller networks to predict the next state and reward independently.  This was because the original algorithm led to unstable training and poor convergence, perhaps due to the problem's high dimensionality. In practice, our implementation led to more stable and quicker training of the agent.
    
    \item \textbf{\texttt{buffer.py:}} Class definition for \texttt{ReplayBuffer}, which is required to implement the experience replay mechanism, first introduced by \citeauthor{Mnih2015HumanlevelCT} in 2015.
    
    \item \textbf{\texttt{ddpg\_agent.py:}} Class definition for \texttt{DDPGAgent}, which utilises the \texttt{model} and \texttt{buffer} modules to abstract the implementation of the Dyna-DDPG algorithm. The backbone of this class is the \texttt{update\_actor\_network} method, which updates the actor network following the Dyna-DDPG algorithm, the \texttt{update\_critic\_network} method which updates the critic network following the Dyna-DDPG algorithm, and the \texttt{fit\_model} method which trains the agent's internal model of the environment (next state and reward predictor networks) using all experiences within its replay buffer.
\end{itemize}

\subsubsection{Queueing Environment}
The \texttt{queue\_env} module is responsible for the construction of a queueing network object based on a provided configuration. It is organized into three major sections, each with specific functionalities:

\begin{enumerate}
    \item \textbf{\texttt{queueing\_network.py}:} This file contains the \texttt{Queue\_network} class, which is used to instantiate a queueing environment.
    
    \item \textbf{\texttt{queue\_base\_functions.py}:} This script provides support functions essential for setting up the queueing network class. It details major functions and attributes necessary for its operation.
    
    \item \textbf{\texttt{queueing\_foundations}} folder: This directory is a modified version of the open-source GitHub repository for the queueing tool package. For comprehensive documentation of its original implementation, please refer to \href{https://queueing-tool.readthedocs.io/en/latest/}{queueing-tool documentation}. We have included this folder in our repository because we have customized certain functions and class attributes within \texttt{queue\_server.py} to accommodate specific simulation needs. Consequently, users are required to install this locally adjusted version of the queueing tool to function properly with our enhancements.
\end{enumerate}

\bigskip
In modelling the queueing networks, the following assumptions were made: 
\begin{itemize}
    \item There is no limit to the number of jobs that can wait in the queue. The queue operates on a first-in, first-out basis with jobs being served in the order they arrive. Therefore, all jobs have been treated equally without differing levels of priority that could affect their service order. 
    \item Other assumptions have also been made around traffic intensity, where the average arrival rate is designed to be less than the average service rate such that the queue does not grow indefinitely. And jobs are assumed to behave rationally, following the agent’s decision without deviating.
    \item Finally, it was assumed that in the real life system, all queues going into a server had the same service rate as they would all undergo the same process. Service rates were all taken to be deterministic as it was assumed that machines in the network had negligible changes in performance between jobs. 
\end{itemize}

\subsubsection{RL Environment}
The \texttt{RLEnv} class serves as the simulation environment for training agents within our framework, employing the \texttt{QueueingNetwork} class to construct an operational representation for agent interactions. Key functionalities of the \texttt{RLEnv} class include the \texttt{get\_next\_state} method, which takes an action vector from the actor network and converts it to a transition probability representation, and then uses the \texttt{simulate} function to process jobs through the network accordingly. The \texttt{get\_reward} method computes the reward for a given state, assisting in the evaluation of agent actions, while the \texttt{get\_state} method provides the current state of the environment, ensuring up-to-date feedback for the training process. Collectively, these methods enable the RL agent to retrieve a simulated experience in the queueing environment. 

\subsubsection{Blockage Exploration}
To make the RL agent learn more adaptive policies reflective of real-world scenarios, we introduced \linebreak \texttt{breakdown\_exploration} module. In this approach, a single server in the network is rendered non-functional by setting its service rate to infinity. Training the agent to account for such states enables it to dynamically adapt to environmental changes in deployment. Moreover, to balance the exploitation-exploration trade off, a weighted approach to exploring key and peripheral states is employed to determine the starting state for subsequent episodes. To explain, key states refer to the states that have significant impact on the rewards received by the agent, whereas peripheral states record states according to the number of visits by the agent regardless of their reward. Tracking key states is important to optimize the effectiveness and efficiency of the agent, while including peripheral states is important to enhance the agent’s exploration behaviour. Thus, this strategy balances exploration and exploitation trade-off ensuring a comprehensive understanding of the environment and a robust decision-making process.

\subsection{Hyper-parameter tuning and evaluation procedures}
As our overall approach went through numerous iterations from altering the agent, the environment, and reward function it proved troublesome to manually find the best hyper-parameters for the given code. Therefore, a Yaml file was created for the user to easily insert the values to trial. The chosen hyper-parameters\footnote{Refer to Appendix A for a list of hyperparameters (page \pageref{appendix:list of hyperparameters})} to vary were selected based on balancing run time and coverage with values and ranges have been extensively researched \citep{ mathworks2024rlddpgagentoptions, haijun2021iot}.The  \texttt{tuning} module offers two hyperparameter tuning frameworks, \texttt{Wandb} and \texttt{Ray Tune}.\par 

\subsubsection{RayTune}
The main method within the \texttt{ray\_tuning.py} file is \texttt{ray\_tune()}, utilizing the \texttt{PopulationBasedTraining} class for conducting a multi-parallel search. It draws from a set of predefined hyperparameter mutations, with defaults set for 10 parallel agents, a 0.120-second perturbation interval, and a 25\% resampling probability, though these can be customized by the user. Agents with varying hyperparameters are trained independently in parallel, assessed by the \texttt{perturbation\_interval}. The best models influence the next training cycle’s population through hyperparameter adoption, with the \texttt{resample\_probability} metric enabling resampling from a distribution. Training outcomes and optimal parameters are displayed in the terminal, and logging information can be accessed from \texttt{ray\_tune()} results.

\subsubsection{WandbTune}
The \texttt{wandb\_tuning.py} file's main function is \texttt{wandb\_train} offered by Weights \& Biases, a Machine Learning development platform that allows users to track and visualize various aspects of their model
training process in real-time \citep{educative2024wandb}. After reading the YAML file, the tuning process starts by initializing \texttt{Wandb} for project tracking and setting up a hyperparameter tuning sweep. The process is automated through a sweep function in \texttt{Wandb}, optimizing performance by running multiple experiments to find the best hyperparameter combination. The training is initiated with the RL environment and the agent set with these configurations, with each training phase logged to \texttt{Wandb}.

\subsubsection{Evaluation Functions}
Throughout the development lifecycle of the agent, different modules were concurrently created to continually assess distinct dimensions of agent performance and robustness. Follows are the five modules used in evaluating the agents learning:

\begin{itemize}
    \item The \textbf{startup\_evaluation} is focused on identifying the end of the burn-in period of an agent during training, pinpointing when the agent surpasses its startup behavior and begins to learn stably.
    \item The \textbf{convergence\_evaluation} includes a modified training function that periodically reviews the agent's progress, stopping the training when the policy has sufficiently converged.
    \item The \textbf{noise\_evaluation} assesses how well the learned policy withstands noise introduced into the evaluation setting. Specifically, testing how the agent would adapt to out-of-distribution inter-arrival times.
    \item The \textbf{decision\_evaluation} section utilizes the \textbf{DisruptionEvaluation} class to examine a pre-trained agent’s performance in environments with user-specified blockages. This module is crucial for analyzing changes in transition probabilities and throughput before and after the disruption.
    \item The \textbf{robustness\_evaluation} module encompasses the \textbf{RobustnessEvaluation} class, which calculates the variance in the final decisions made by a predefined group of pre-trained agents. The measurement of variance is used to estimate the necessary number of runs required to achieve the desired robustness within a specified confidence interval and error rate. This provides an evaluation of the agent’s reliability and consistency in decision-making under simulated conditions.
\end{itemize}

\subsection{Software engineering practices}
\subsubsection{Version Consistency}
The \texttt{Poetry} package was utilized for version control to maintain systematic updates and configuration changes across all team members. Specifically, the \texttt{pyproject.toml} file tracked all package and development dependencies. Each update to the \texttt{pyproject.toml} file prompted an update to the poetry.lock file using the \texttt{poetry lock command}. The poetry.lock file ensured that all package versions are consistently installed across all team members' environments. The virtual environment was created using \texttt{poetry install}. Throughout the development process, commands such as \texttt{poetry add} and \texttt{poetry remove} were employed to update the dependencies as necessary.
The use of \texttt{poetry} helped enhance dependency resolution and package management, creating reproducible builds and streamlining the update process. Finally, \texttt{poetry build} and \texttt{poetry publish} were used to publish our pypl package, \href{https://pypi.org/project/sim_rl/}{sim\_rl}. 

\subsubsection{Code Formatting}
The code was formatted using \texttt{black} to maintain a consistent coding style throughout the package. General coding best practices governing, function/class names, comment inclusion and the use of docstrings were adhered to throughout the repository. 
\section*{Analysis}
\label{section:analyis-and-evaluation}
\begin{figure}[h]
    \centering
    \vspace{-1ex}
    \includegraphics[width=0.5\textwidth]{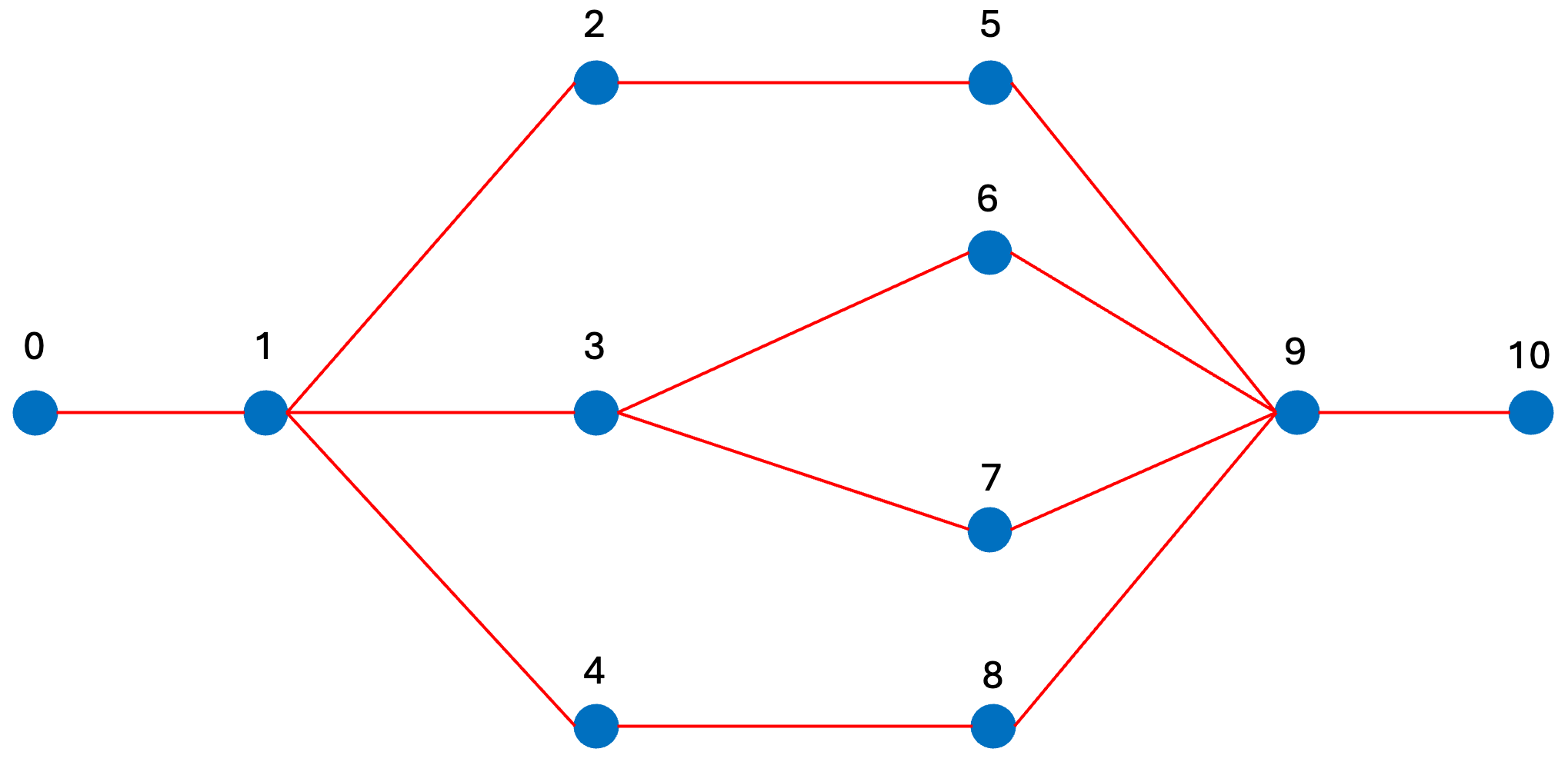}
    \caption{Queueing network used for analysis}
    \label{fig:network_for_analysis}
\end{figure}

Figure \ref{fig:network_for_analysis} shows the standard queueing network which was analysed using the developed package. There are 11 nodes in total, with a single entry and exit node. The questions that follow were posed by Datasparq and encapsulate the core considerations for comprehensive evaluation of the learned policy. These were used to guide the development of our evaluation suite. In response to each question, we have detailed our chosen approach for evaluation and the corresponding analysis with respect to the given configuration. 

\subsection{How do we remove or discount start-up behaviour in the simulation?}
\begin{figure}[h!]
    \centering
    \includegraphics[width=0.7\linewidth]{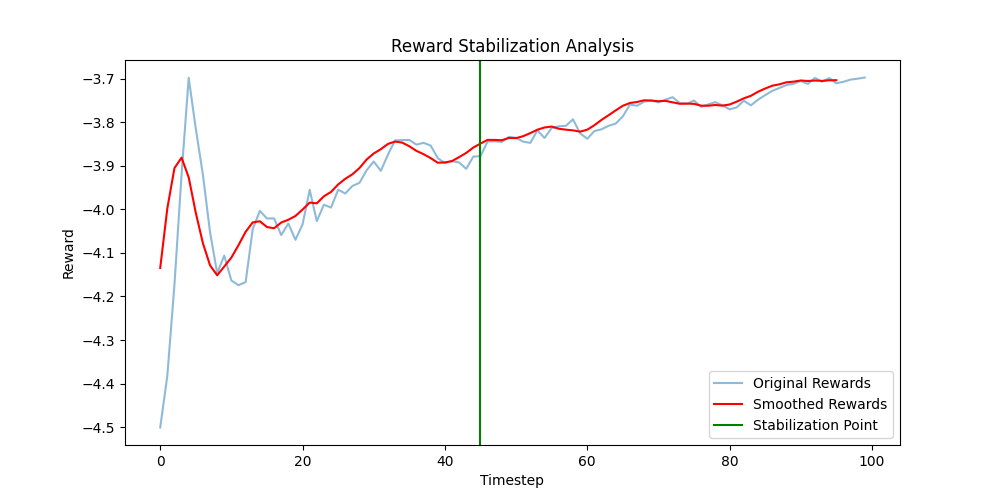}
    \vspace{-1em}
    \caption{Reward start-up behaviour}
    \label{fig:reward_stabilisation_plot}
    \vspace{-1em}
\end{figure}

In the given RL environment, the actions are initialized as equal transition probabilities to all possible next nodes. This initialisation causes the agent to be unstable at the beginning of training and it may take some time to converge to a stable state. Addressing the start-up behavior --- also known as the `burn-in period' --- is achieved by the \texttt{StartupBehavior} class. This class tracks the reward trajectory per timestep for a particular episode, by applying a moving average to smooth transient fluctuations and then calculating the derivative to pinpoint stabilisation. By defining a stabilisation threshold and requiring a sequence of consecutive data points to fall beneath this threshold, it identifies when the agent’s performance has transitioned from exploratory to consistent. The plot in Figure~\ref{fig:reward_stabilisation_plot} shows the time step where the reward has stabilised as indicated by the green line.

\subsection{Should we run many instances of the simulation to test a single decision?}
\label{subsection:analysis-q2}

To ensure that the final policy is robust and has low uncertainty around its final decisions, it is essential to run multiple instances of training. However, this requires substantial computational resources. To determine the minimum number of instances needed to achieve a specified level of robustness, we have introduced the \texttt{robustness\_evaluation} module. This module performs the following steps:

\begin{itemize}
    \item \textbf{Training}: Trains multiple agents (default is 10) using the settings specified in \linebreak \texttt{user\_config/eval\_hyperparameters.yml}.
    \item \textbf{Simulation}: Each agent executes a predetermined number of actions (\texttt{time\_steps}) in a simulated environment that maintains the same initial conditions for consistency.
    \item \textbf{Evaluation}: The maximum standard deviation in the final transition probabilities is used as a conservative estimate of an agent's confidence in its decisions.
\end{itemize}

\begin{figure}[h!]
    \centering
    \includegraphics[width=0.48\linewidth]{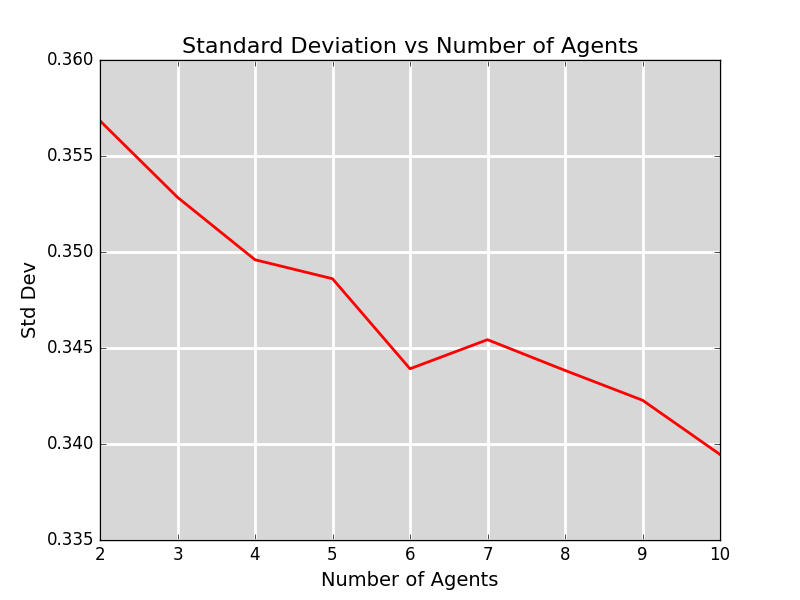}
    \caption{Standard deviation against number of trained agents}
    \label{fig:Std}
    \vspace{-1em}
\end{figure}

\bigskip
The necessary sample size is then calculated based on a user-defined margin of error ($E$) and confidence level ($z$-score) using the following equation: 
\[
n = \left( \frac{z \times \sigma}{E} \right)^2
\]
Figure~\ref{fig:Std} shows that, as expected, the variability in the agents final decisions decreases as the number of agents increase. It is also worth noting that the standard deviation remain in a consistent range across the different number of agents indicating that the diversity in the decisions does not vary widely and running more instances would not lead to a proportional decrease in the variability of decisions. Given this range of the standard deviation and a confidence level of 95\% indicates a single agent is sufficient given the margin of error ($E$) is equal to 1. 

\subsection{How much learning coverage should we get of key states or peripheral states in our system?}

On one hand, key states are those that are frequently visited and critical for the success of the policy being learned. Since operating under normal conditions is the most common scenario and exercise a great impact on the policy's development, we define the key states in our project as the usual conditions where there are no outages occur in the queues. Therefore, it is essential to ensure agent learning performance under normal conditions. To achieve this, we provide aforementioned tuning files \texttt{wandb\_tune.py} and \texttt{ray\_tune.py} to allow the agent to find the best hyperparameters to perform optimally in a given scenario.

\bigskip
On the other hand, peripheral states are those that occur less frequently or being less visited by the agents. In real-world situations, they may represent rare but crucial servers being broken down that can deteriorate system’s performance. Therefore, we enforce the agent to expose under various breakdown scenarios, such that it can perform well enough to handle unexpected outages. 

\bigskip
To address the problem, we introduce blockage exploration to train our agent with various servers being set broken manually, such that it can achieve high rewards under both blockage and non-blockage cases. However, to balance the agent's training between these two cases, we introduce weight parameters, as discussed below.

\bigskip
To determine the learning coverage between key states and peripheral states, we introduce \textit{w1} and \textit{w2} respectively. A high value of \textit{w1} means the agent will be trained more often with normal scenarios, while a high value of \textit{w2} means the agent will explore blockage cases more. A parameter value of 0.5 for each means an equal learning coverage, which has proven to perform well under outage scenario when the target node 3 connecting to Queue 2 is set to be broken, as shown below.

\begin{figure}[ht!]
  \centering
  \begin{minipage}[b]{0.31\textwidth}
    \includegraphics[width=\textwidth]{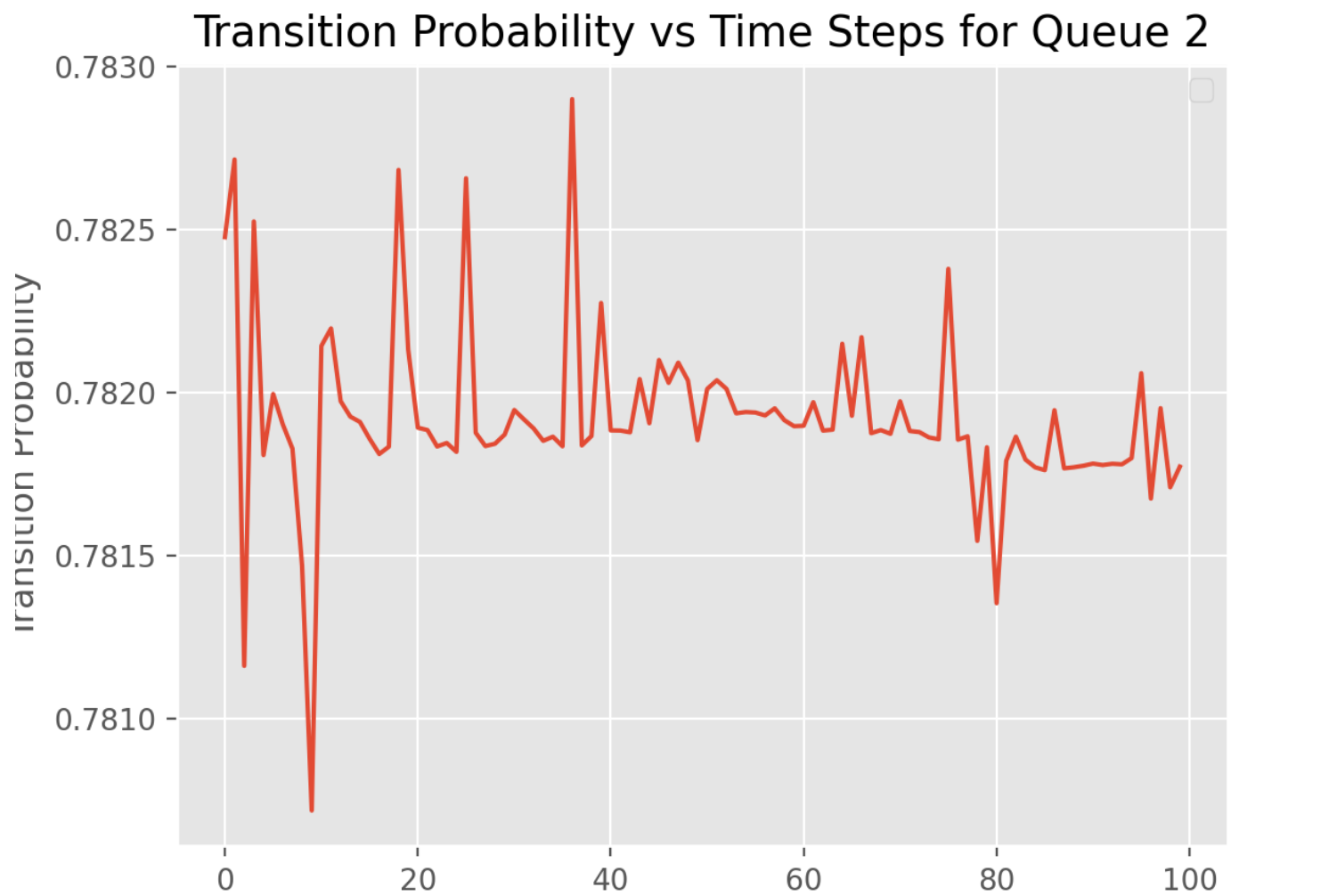}
    \caption{Before blocking}
    \label{fig:tran_proba_before}
  \end{minipage}
  \hfill 
  \begin{minipage}[b]{0.31\textwidth}
    \includegraphics[width=\textwidth]{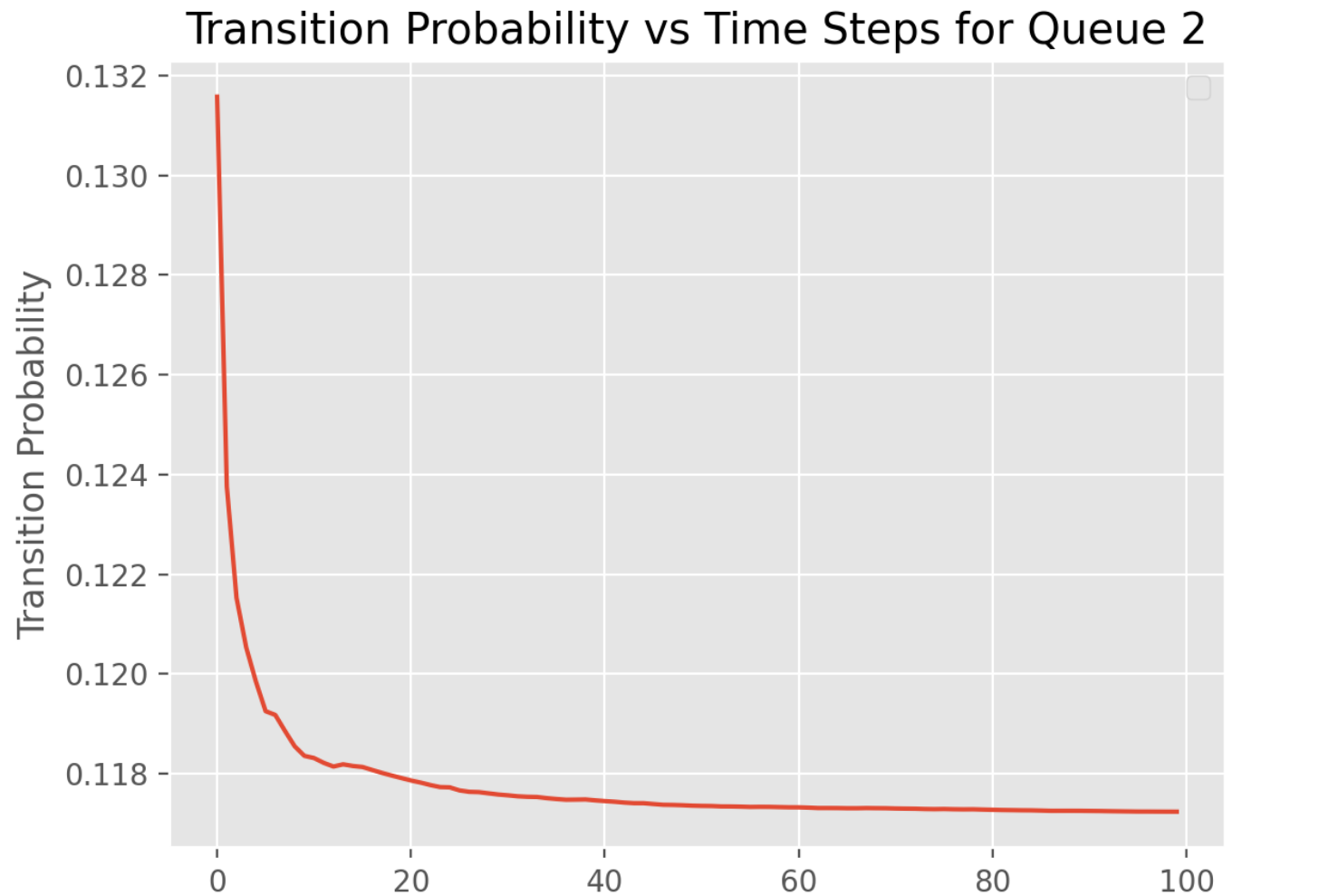}
    \caption{After blocking}
    \label{fig:tran_proba_after}
  \end{minipage}
  \hfill 
  \begin{minipage}[b]{0.31\textwidth}
    \includegraphics[width=\textwidth]{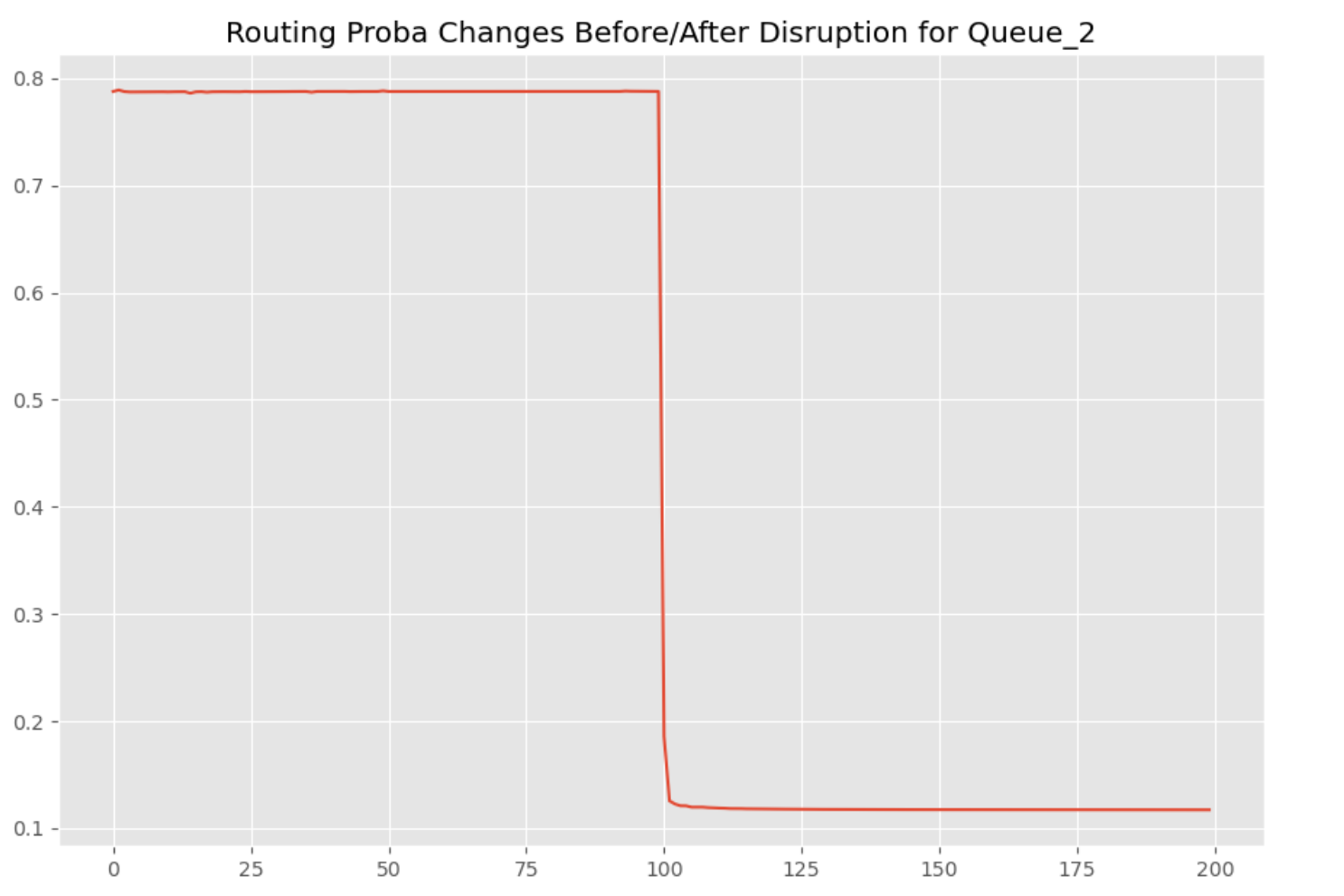}
    \caption{Combined plot}
    \label{fig:tran_proba_before_after}
  \end{minipage}
\end{figure}

In the plots, \(x\)-axis represents the time steps and \(y\)-axis is the routing probability for Queue 2. As we can see, the routing probability allocated to Queue 2 before it is being blocked hovers around 0.782. After the agent being trained on various blockage scenarios, this probability significantly drops to only around 0.05. As a result, training under outage cases can enable the agent to deal with blockages effectively. 

\bigskip
However, there are some limitations in our blockage exploration. We only expose the agent to scenarios where only one server is being blocked and does not account for scenarios where multiple servers being blocked. Therefore, future studies are needed to shed light on this problem.

\subsection{Are there any rules of thumb for the level of confidence we can have in the RL learner (optimality of decision) versus the amount of learning or simulation time we need to run our simulation-learner system for?}
\noindent  

The \texttt{convergence\_evaluation.py} file was created to evaluate this question. Initially, this code, took a list containing a sequence of episode lengths for the agent to be trained on. At the end of training for one of the listed duration, the agent would then be evaluated on the environment where after 100 time-steps, the final reward would be stored. After this the agent was then retrained from the beginning for the next item in the list and the process repeated. The trained and evaluation environment was the same as those listed before. 

\begin{figure}[!h]
    \centering
    \vspace{-1em}
    \includegraphics[width=0.48\linewidth]{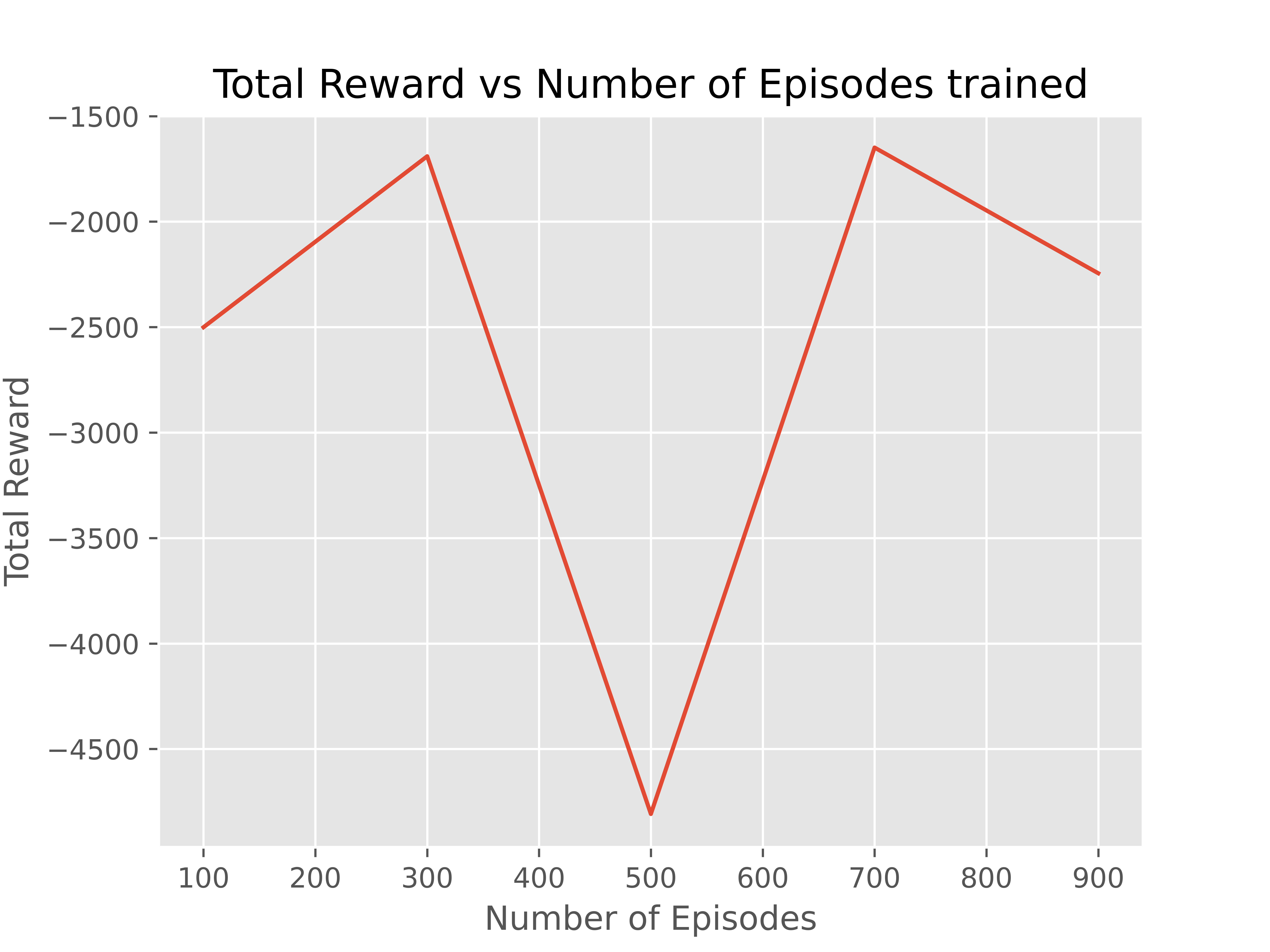}
    \caption{Evaluated reward against number of episodes trained for}
    \label{confidence_plot}
    \vspace{-1em}
\end{figure}

\bigskip
From Figure \ref{confidence_plot} we can break down the shape of the graph into 4 main sections:
\begin{enumerate}[leftmargin=*]
    \item From 100 to 300 episodes, the agent experiences improvements in learning. This could be due to the agent exploring the environment, discovering more efficient routing proportions, and converging towards a better policy. As a result, the total reward of the system increases from $-2500$ to $-1750$.
    \item However, as the agent continues training beyond 300 episodes and approaches 500 episodes, it might start overfitting to the training environment or getting stuck in local optima. This could lead to a sudden drop in performance, where the agent's learned policy becomes less effective in the evaluation environment. Thus, the total reward of the system dramatically decreases to $-4750$.
    \item After training beyond 500 episodes, the agent might explore alternative policies or continue refining its current policy. This exploration phase could lead to the discovery of better routing proportions, resulting in an improvement in the total reward of the system. Hence, we observe a rise back to $-1750$ at 700 episodes.
    \item From continuing training beyond 700 episodes, the agent may experience overfitting to the training environment. This could result in a drop in performance once again, leading to the total reward of the system decreasing at 900 episodes. 
\end{enumerate}

\bigskip
Overall, our recommendation would most likely be training the agent for around 300 episodes on smaller systems, and increasing episodes as the system obtains more complexity (i.e. more nodes or looped nodes, etc).

\bigskip
When executing the code for this graph, it was found that the execution time was extremely long, taking over 30 hours to complete the training and evaluation of the 5 points. As a result, this file was reworked with a focus on improving efficiency. A similar method to early stopping was implemented and takes 5 user defined variables:

\bigskip
\begin{enumerate}
    \item \texttt{window\_size}: Size of window to observe when calculating moving average
    \item \texttt{threshold}: Minimum derivative difference in order to be considered a change
    \item \texttt{agent}: The path to the agent's configuration file
    \item \texttt{consecutive\_points}:  Number of consecutive different points before early stopping
    \item \texttt{env}: The training environment file path 
\end{enumerate}

\bigskip
After inputting these variables and running the file, the code will initialise the \texttt{ConvergenceEvaluation} class and \texttt{start\_train} function will be executed. An agent will be trained and evaluated for the given window size. At each interval of 10 episodes, the agent's current learning will be evaluated on the system and the current reward is stored. The class uses a method of early stopping, such that after the set number of consecutive reward points decrease from each other surpassing the difference threshold this indicates that a local maxima has been reached and thus the program will stop. Alternatively if the reward shows no difference for the set number of consecutive points the evaluation will also be stopped as this indicates a plateau in reward vs episode. The total reward of the agent at every 10 episodes is stored and plotted after all training episodes have been evaluated.  The resulting plot is saved under 
 \texttt{reward\_plot.png} in the \texttt{evaluation/convergence\_evaluation} folder. It was found that, this new method resulted in a significant improvement in efficiency.

\subsection{How does noise in the simulation obstruct or interfere with learning?}

In reality, interarrival rates are stochastic and may not consistently adhere to a specific distribution. To address this variability, we have developed the \texttt{noise\_evaluation} model, which trains and evaluates agents in a noisy network environment. The model introduces the \texttt{NoisyNetwork} class, responsible for generating a noisy environment by adding increments to the interarrival time, thus altering its original predefined distribution. These increments are randomly drawn from a normal distribution with a user-specified \texttt{mean} and \texttt{variance} (default is mean 0 and variance 1) and are applied to the arrival rate based on a defined probability --- the \texttt{frequency} input to the class.
\bigskip

\begin{figure}[h!]
    \centering
    \includegraphics[width=0.48\linewidth]{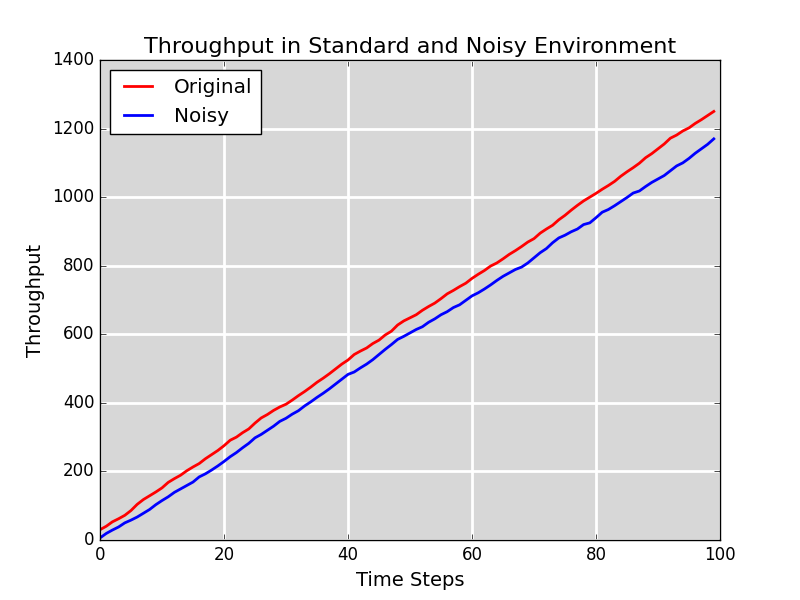}
    \caption{Throughput rate in standard and noisy environments}
    \label{noise_plot}
    \vspace{-1em}
\end{figure}

During each simulation step, there is a 50\% chance (default probability) that the interarrival time will be perturbed by this increment. The \texttt{NoisyNetwork} class primarily serves two functions: retraining the agent within this noisy environment or evaluating a pre-trained agent’s performance under new, noisy conditions. As illustrated in Figure~\ref{noise_plot}, although introducing noise during evaluation slightly reduces the throughput rate, the overall trend of increasing throughput persists. This indicates that the agent’s policy successfully adapts to maximize rewards, even with the introduced noise.

\newpage 
\section*{Evaluation}
\subsection{Learning Assessment and Visualization}
To visualize agent learning, three main functions were used:
\begin{itemize}
    \item \texttt{plot\_transition\_proba()}
    \item \texttt{plot\_reward()}
    \item \texttt{plot\_average\_reward\_episode()}
\end{itemize}

\bigskip
The \texttt{plot\_transition\_proba()} function is used to visualize how the agents learn to traverse the action space for any specified node. By default, agents start with equal transition probability to all possible next nodes for a given source node. However, as the agent learns the optimal policy, it learns to assign higher transition probability for the next node path of the highest reward, the path with the lowest end-to-end delay, and highest throughput ratio. In the example below, using a simple configuration, where node 1 is connected to three different nodes; 2, 3, and 4. And the processing time of each node is given as exponential distribution of 0.25, 0.0015, and 100, respectively. The plot shows the agent starts from equal transition probabilities but eventually learns and stabilizes at assigning the highest probability to node 3 of the fastest service rate, and the lowest probability to node 4 of the slowest processing time.

\begin{figure}[h]
\centering
\vspace{-2em}
\includegraphics[width=0.6\textwidth]{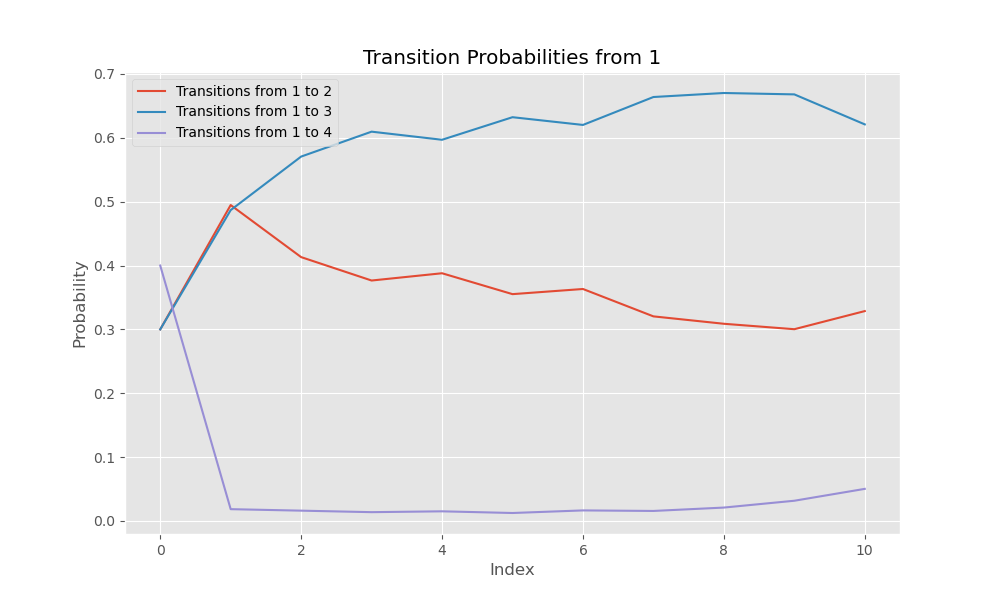}
\vspace{-1em}
\caption{Transition probabilities from node 1}
\label{fig:tran_proba}
\vspace{-1em}
\end{figure}

\begin{figure}[ht]
  \centering
  \begin{minipage}{0.48\textwidth}
    \centering
    \includegraphics[width=\linewidth]{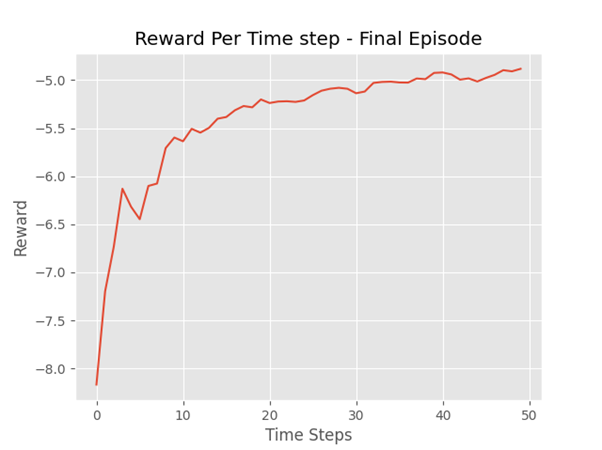}
    \caption{Reward per time step}
    \label{fig:Reward Per Time Step}
  \end{minipage}
  \hfill 
  \begin{minipage}{0.48\textwidth}
    \centering
    \includegraphics[width=\linewidth]{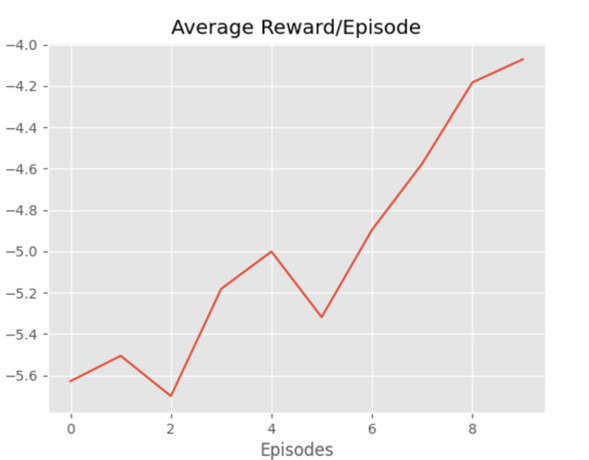}
    \caption{Average reward per episode}
    \label{fig:Average Reward Per Episode}
  \end{minipage}
  \vspace{-1em}
\end{figure}

\bigskip
The \texttt{plot\_reward()} function is used to demonstrate the agent’s reward per time step for a given episode --- the last episode by default. This plot shows how the agent's learning progresses during an episode. As the episode progresses, the agent learns from an increasing number of experiences, thus, we expect the reward to increase and converge at an optimal value, as shown in Figure~\ref{fig:Reward Per Time Step}.

\bigskip
The \texttt{plot\_average\_reward\_episode()} function is used to demonstrate the agents learning curve. It plots the average reward accumulated per episode. As training progresses, the agent should be starting the episode with an improved policy compared to previous episodes. Hence, the learning curve is expected to start at a low reward value and increase to the reward yielded by the optimal policy. It is also worth noting that the reward curves yielded from \texttt{plot\_reward()} and 
\texttt{plot\_average\_reward\_episode()} are used to identify the convergence (stabilisation) points in terms of the required number of time steps and episodes. This aids in implementing early stopping once training begins to converge.

\bigskip
In addition to the three main functions, the following functions are used to demonstrate different models’ loss changes during training:

\bigskip
\begin{itemize}
    \item \texttt{plot\_actor\_loss()}
    \item \texttt{plot\_critic\_loss()}
    \item \texttt{plot\_next\_state\_model\_loss()}
    \item \texttt{plot\_reward\_model\_loss()}
\end{itemize}

\begin{figure}[ht]
  \centering
  \begin{minipage}{0.48\textwidth}
    \centering
    \includegraphics[width=\linewidth]{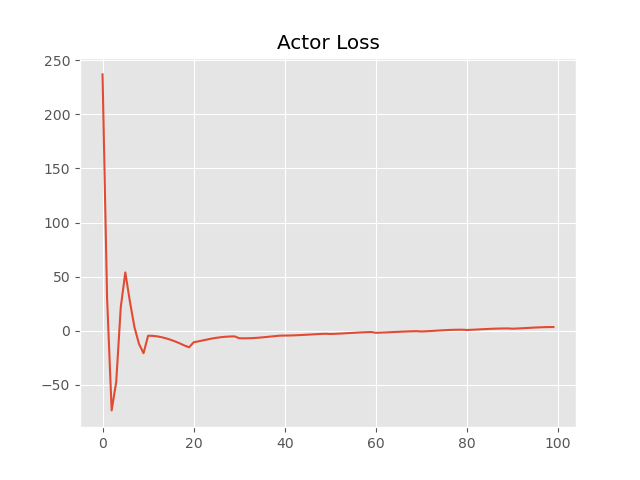}
    \vspace{-2em}
    \caption{Actor loss}
    \label{fig:ActorLoss}
  \end{minipage}
  \hfill 
  \begin{minipage}{0.48\textwidth}
    \centering
    \includegraphics[width=\linewidth]{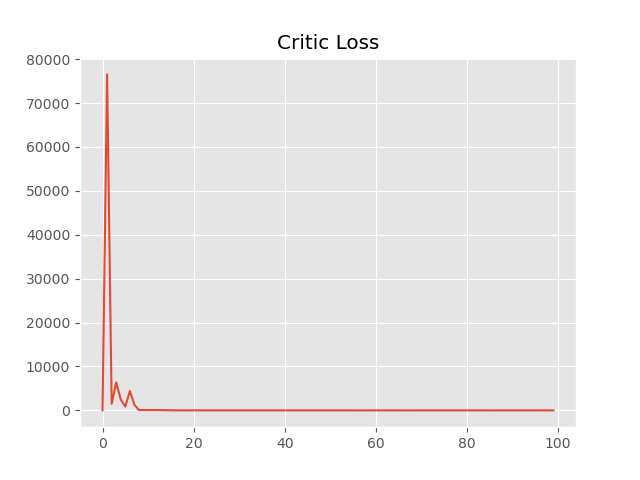}
    \vspace{-2em}
    \caption{Critic loss}
    \label{fig:CriticLoss}
  \end{minipage}
\end{figure}

\bigskip
The function \texttt{plot\_actor\_loss()} is used to track the improvement of the policy (the actor) throughout training. The actor network selects actions based on the current state, aiming to maximize the probability of choosing actions that yield the highest expected rewards. The actor's loss is calculated as the negative of the expected return. Specifically, this involves taking the negative of the critic's output (Q values) for the state and action selected by the actor. By minimizing its loss, the actor effectively maximizes these Q-values. As depicted in Figure~\ref{fig:ActorLoss}, the actor initially experiences a high loss, which gradually stabilizes and converges to a lower policy loss, indicating that the agent is learning to maximize the Q values effectively.

\bigskip
The \texttt{plot\_critic\_loss()} is given as the mean squared error between the predicted Q-value and the actual Q- value received. The predicted value is derived from the target critic network, while the actual Q-value is based on the value of the current policy derived from the actor network. A decreasing trend is expected in critic’s loss to show that the critic network is learning to accurately predict future rewards. The \texttt{plot\_next\_state\_model\_loss()} and \texttt{plot\_reward\_model\_loss()} represent the loss associated with models used to compute the next state and the reward. These models are used to simulate experiences in Dyna planning. The lower loss indicates the models are getting better at predicting next states and rewards given a state and an action.
\begin{figure}[H]
  \centering
  \begin{minipage}{0.48\textwidth}
    \centering
    \includegraphics[width=\linewidth]{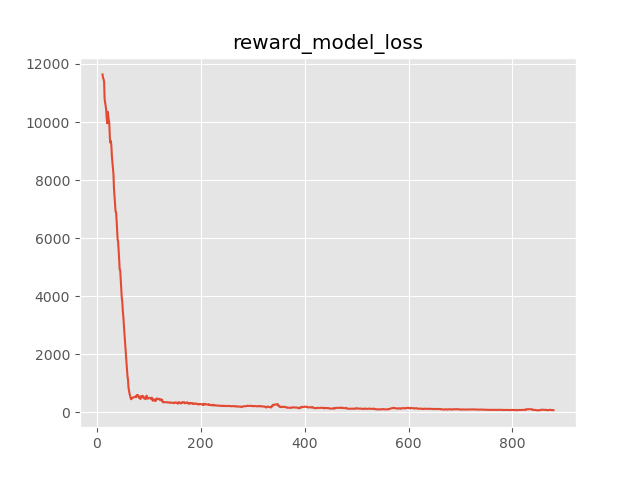}
    \vspace{-2em}
    \caption{Reward Model loss}
    \label{fig:RewardModelLoss}
  \end{minipage}
  \hfill 
  \begin{minipage}{0.48\textwidth}
    \centering
    \includegraphics[width=\linewidth]{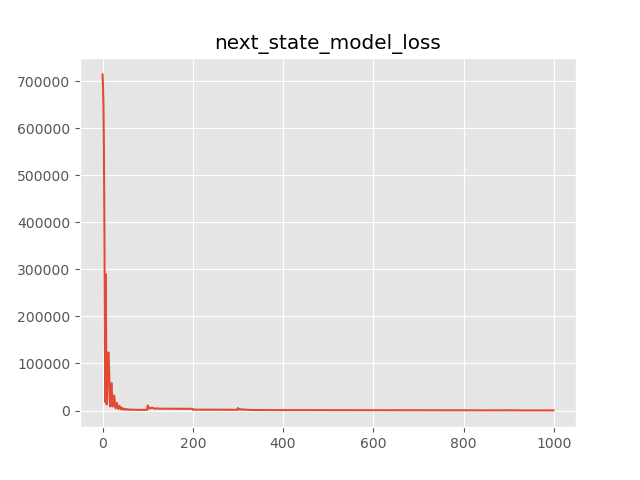}
    \vspace{-2em}
    \caption{Next Model loss}
    \label{fig:NextModelLoss}
  \end{minipage}
\end{figure}

\subsection{Performance Scaling}
The script \texttt{robustness\_evaluation.py} (section \ref{subsection:analysis-q2}) was also used to assess how well our Dyna-DDPG training algorithm scales with different network sizes. The maximum standard deviation which estimates a trained agent's confidence in its actions is also a measure of performance (robustness), as a poorer-performing agent is expected to have less certainty in what action to take given a state. Hence, it was reasonable to extend the functionality of this script to performance scaling. The results show that one instance (run) was sufficient for a network with 10 nodes, but two instances were necessary for networks with 50 and 100 nodes. Thus, the algorithm scales well for networks up to 100 nodes. Due to compute constraints, larger configurations were not tested.



\subsection{Testing}
Our project employed comprehensive testings to ensure that our software remains robust during every stage of development. Specifically, our testings include unit tests for individual components, such as agents and environments initialization, integration tests to ensure interactions between these components, and system abnormality tests to successfully simulate a blocked environment. 

\bigskip
Throughout the development process, the code was tested continually using the \texttt{pytest}, a testing package, to ensure that the software remained robust during production. These can be broadly be categorized into unit tests for individual component functionality and integration tests to ensure compatibility between the respective objects. The team ensured were passed upon merging feature branches with the main development branch.

\bigskip
In total 48 tests were written --- the respective details are specified in each function's docstrings, following the pattern illustrated below:

\bigskip
\begin{itemize}
    \item \texttt{test\_agents.py:} Rigorously evaluates the functionalities of DDPG (Deep Deterministic Policy Gradient) agents, focusing on initialization, action selection, and learning processes.
    \item \texttt{test\_buffer.py:} Addresses the Replay Buffer’s operations, crucial for managing the historical data that agents use for learning. This suite ensures that the buffer efficiently manages its capacity and handles scenarios such as overflows and underflows gracefully.
    \item \texttt{test\_model.py:} Assesses the integrity and accuracy of neural network models such as Actor and Critic, ensuring that they meet expected output standards and handle inputs correctly.
    \item\texttt{test\_queueing\_network.py} and \texttt{test\_RL\_Environment.py}, which verify system configurations and simulate dynamic interactions within the RL environment, ensuring that the system accurately reflects complex real-world scenarios.
\end{itemize}
\newpage
\section*{Conclusions}

In this project, we developed a simulation-based reinforcement learning (RL) framework, tailored specifically for optimising the routing problem presented by network systems. Our particular use case focused on manufacturing systems; however, this library can be extended to any network problem that can be modelled using queueing theory. The library provides a suite of evaluation tools, allowing users to thoroughly analyse agent performance before deployment into live systems.

\bigskip
This library provides a solid foundation upon which further research can be performed. We have identified the following areas as those which present the most crucial scope for additional development.

\begin{itemize}
    \item System Configuration - The current functionality of this package can be extended to encompass more complex route optimisation scenarios such as those that consider loop re-routing proportions and dynamic server allocation.
    \item Model Complexity - Our framework can be adapted to models of greater complexity which feature non-Markovian distributions of interarrival and service times and varying queueing schedules.
    \item Learning Algorithm - Our chosen reward function and state representation approach produced promising results; however, by incorporating additional networking expertise, these could potentially be refined to achieve enhanced learning outcomes. Furthermore, the evaluation of alternative deep reinforcement learning algorithms, such as the Soft Actor-Critic method, could provide additional insights. Lastly, to facilitate the exploration of more complex configurations, the package can be extended to support multi-agent reinforcement learning.
\end{itemize}

\section*{Appendices}
\subsection*{Appendix A: List of hyperparameters}
\label{appendix:list of hyperparameters}
The list of agent hyperparameters is as follows:
\begin{enumerate}
    \item Learning rate ($\alpha$)
    \item Number of epochs
    \item Batch size
    \item Number of planning steps
    \item Number of samples
    \item Number of episodes
    \item Number of time steps
    \item Target update frequency
    \item Tau ($\tau$) - soft update parameter for target networks
    \item Future rewards discount 
    \item Epsilon ($\varepsilon$) - noise for action exploration in planning
    \item \textit{w1} - weights for key state 
    \item \textit{w2} - weights for peripheral state
\end{enumerate}

\subsection*{Appendix B: Package installation and usage}
\label{appendix:installation and usage}



\textbf{Installation}. To get started with this project, clone the repository and install the required dependencies as follows:

\begin{verbatim}
git clone https://github.com/ao-420/sim_rl.git
cd sim_rl
pip install -r requirements.txt
\end{verbatim}

\textbf{Usage}. The repository's README file has detailed instructions on how to train the agent, tune hyperparameters, and utilize the evaluation features. For example, to train the agent, use:

\begin{verbatim}
python main.py \
    --function train \
    --config_file user_config/configuration.yml \
    --param_file user_config/eval_hyperparams.yml \
    --data_file output_csv \
    --image_file output_plots \
    --plot_curves True \
    --save_file True
\end{verbatim}

\begingroup
\newpage
\singlespacing
\bibliographystyle{agsm}
\bibliography{references.bib}

@misc{zhang2020deep,
      title={Deep Residual Reinforcement Learning}, 
      author={Shangtong Zhang and Wendelin Boehmer and Shimon Whiteson},
      year={2020},
      eprint={1905.01072},
      archivePrefix={arXiv},
      primaryClass={cs.LG},
}

@misc{lillicrap2015continuous,
      title={Continuous control with deep reinforcement learning}, 
      author={Timothy P. Lillicrap and Jonathan J. Hunt and Alexander Pritzel and Nicolas Heess and Tom Erez and Yuval Tassa and David Silver and Daan Wierstra},
      year={2015},
      eprint={1509.02971},
      archivePrefix={arXiv},
      primaryClass={cs.LG}
}

@article{Mnih2015HumanlevelCT,
  title={Human-level control through deep reinforcement learning},
  author={Volodymyr Mnih and Koray Kavukcuoglu and David Silver and Andrei A. Rusu and Joel Veness and Marc G. Bellemare and Alex Graves and Martin A. Riedmiller and Andreas Kirkeby Fidjeland and Georg Ostrovski and Stig Petersen and Charlie Beattie and Amir Sadik and Ioannis Antonoglou and Helen King and Dharshan Kumaran and Daan Wierstra and Shane Legg and Demis Hassabis},
  journal={Nature},
  year={2015},
  volume={518},
  pages={529-533},
  url={https://api.semanticscholar.org/CorpusID:205242740}
}

@misc{shen2020deep,
      title={Deep Reinforcement Learning with Robust and Smooth Policy}, 
      author={Qianli Shen and Yan Li and Haoming Jiang and Zhaoran Wang and Tuo Zhao},
      year={2020},
      eprint={2003.09534},
      archivePrefix={arXiv},
      primaryClass={cs.LG}
}

@article{dyna1991,
author = {Sutton, Richard S.},
title = {Dyna, an integrated architecture for learning, planning, and reacting},
year = {1991},
issue_date = {Aug. 1991},
publisher = {Association for Computing Machinery},
address = {New York, NY, USA},
volume = {2},
number = {4},
issn = {0163-5719},
url = {https://doi.org/10.1145/122344.122377},
doi = {10.1145/122344.122377},
journal = {SIGART Bull.},
month = {jul},
pages = {160–163},
numpages = {4}
}

@article{Buzacott1986QueueingNM,
  title={On queueing network models of flexible manufacturing systems},
  author={J.A. Buzacott and D.D. Yao},
  journal={Queueing Systems},
  year={1986},
  volume={1},
  pages={5-27},
  url={https://link.springer.com/article/10.1007/BF01149326}
}

@article{Saini2024Queueing,
author = {Saini, Balveer and Singh, Dharamender and Sharma, Dr},
year = {2024},
month = {03},
pages = {256-266},
title = {Exploring the Role of Queueing Theory in Manufacturing: An Analytical Study},
volume = {2},
journal = {International Research Journal on Advanced Engineering and Management (IRJAEM)},
doi = {10.47392/IRJAEM.2024.0039}
}

@misc{educative2024wandb,
  author = {Educative},
  title = {What is Weights \& Biases (wandb) and what is it used for?},
  year = {2024},
  url = {https://www.educative.io/answers/what-is-wandb-and-what-is-it-used-for},
  note = {Accessed: 2024-04-25}
}

@misc{mathworks2024rlddpgagentoptions,
  author = {MathWorks},
  title = {Options for DDPG agent - MATLAB},
  howpublished = {\url{https://www.mathworks.com/help/reinforcement-learning/ref/rl.option.rlddpgagentoptions.html}},
  year = {2024},
  note = {Accessed on April 25, 2024}
}

@article{haijun2021iot,
author = {Liu, Xiangnan and Zhang, Haijun and Long, Keping and Nallanathan, Arumugam and Leung, Victor},
year = {2021},
month = {09},
pages = {1-1},
title = {Deep Dyna-Reinforcement Learning Based on Random Access Control in LEO Satellite IoT Networks},
volume = {PP},
journal = {IEEE Internet of Things Journal},
doi = {10.1109/JIOT.2021.3112907}
}

@article{Chen2018,
author = {Chen, Chen and Tiong, Robert},
year = {2018},
month = {10},
pages = {1-18},
title = {Using queuing theory and simulated annealing to design the facility layout in an AGV-based modular manufacturing system},
volume = {57},
journal = {International Journal of Production Research},
doi = {10.1080/00207543.2018.1533654}
}

@book{sundarapandian2009queueing,
  title={Probability, Statistics and Queueing Theory},
  author={Sundarapandian, V.},
  chapter={7. Queueing Theory},
  publisher={PHI Learning},
  year={2009},
  isbn={8120338448}
}
\endgroup

\end{document}